\documentclass[journal]{IEEEtran}

\makeatletter
\newif\if@restonecol
\makeatother

\usepackage[mathlines,switch]{lineno}
\usepackage{graphicx}
\usepackage{amssymb}
\usepackage{amsmath}
\usepackage{enumerate}
\usepackage{algorithmic}
\usepackage[linesnumbered,ruled]{algorithm2e}
\usepackage[usenames]{color}
\usepackage[colorlinks=true,linkcolor=blue,citecolor=blue,urlcolor=blue]{hyperref}
\usepackage{multirow}
\usepackage{booktabs}
\usepackage{threeparttable}
\usepackage{lineno}

\begin{document}

\title{Learning to Detect Critical Nodes in Sparse Graphs via Feature Importance Awareness}

\author{Xuwei~Tan,
        Yangming~Zhou,~\IEEEmembership{Senior Member,~IEEE,}
        MengChu~Zhou, ~\IEEEmembership{Fellow,~IEEE}
        and Zhang-Hua~Fu
\thanks{This work was supported in part by the National Natural Science Foundation of China under Grants 72371157, 72031007, the Shanghai Pujiang Programme under Grant 23PJC069, and the FDCT (Fundo para o Desenvolvimento das Cienciase da Tecnologia) under Grant 0047/2021/A1. (Corresponding author: \emph{Yangming Zhou})}
\thanks{X.~Tan was with the Sino-US Global Logistics Institute, Antai College of Economics and Management, Shanghai Jiao Tong University, Shanghai 200030, China. He is now with the College of Engineering, The Ohio State University, OH 43210, USA (e-mail: tanxuwei99@gmail.com). The work is done in Shanghai Jiao Tong University as a Research Assistant.}
\thanks{Y.~Zhou is the Data-Driven Management Decision Making Lab, Shanghai Jiao Tong University, Shanghai 200030, China. He is also with the Sino-US Global Logistics Institute, Antai College of Economics and Management, Shanghai Jiao Tong University, Shanghai 200030, China (e-mail: yangming.zhou@sjtu.edu.cn).}
\thanks{M.C.~Zhou is with the Department of Electrical and Computer Engineering, New Jersey Institute of Technology, Newark, NJ 07102, USA (e-mail: zhou@njit.edu).}
\thanks{Z.-H.~Fu is with the Institute of Robotics and Intelligent Manufacturing, The Chinese University of Hong Kong, Shenzhen, Shenzhen 518172, China. He is also with the Shenzhen Institute of Artificial Intelligence and Robotics for Society, Shenzhen 518172, China (e-mail: fuzhanghua@cuhk.edu.cn).}
}

\maketitle

\begin{abstract}
Detecting critical nodes in sparse graphs is important in a variety of application domains, such as network vulnerability assessment, epidemic control, and drug design. The critical node problem (CNP) aims to find a set of critical nodes from a network whose deletion maximally degrades the pairwise connectivity of the residual network. Due to its general NP-hard nature, state-of-the-art CNP solutions are based on heuristic approaches. Domain knowledge and trial-and-error are usually required when designing such approaches, thus consuming considerable effort and time. This work proposes a feature importance-aware graph attention network for node representation and combines it with dueling double deep Q-network to create an end-to-end algorithm to solve CNP for the first time. It does not need any problem-specific knowledge or labeled datasets as required by most of existing methods. Once the model is trained, it can be generalized to cope with various types of CNPs (with different sizes and topological structures) without re-training. Computational experiments on 28 real-world networks show that the proposed method is highly comparable to state-of-the-art methods. It does not require any problem-specific knowledge and, hence, can be applicable to many applications including those impossible ones by using the existing approaches. It can be combined with some local search methods to further improve its solution quality. Extensive comparison results are given to show its effectiveness in solving CNP.
\end{abstract}
\def\abstractname{Note to Practitioners}
\begin{abstract}
This work is motivated by the problems of identifying influential nodes from a sparse graph or network. Various practical applications can be naturally modeled as critical node problems, e.g., finding the most influential stations or airports within a transportation network, identifying a specific number of people to be vaccinated in order to reduce the overall transmissibility of a virus, and reinforcing the protection over some most important nodes to make the electric network more stable. It proposes an effective end-to-end deep learning algorithm to solve the critical node problem. The proposed approach combines feature importance-aware graph attention network with dueling double deep Q-network. Extensive numerical experiments and comparisons show that our proposed algorithm is highly comparable to state-of-the-art algorithms and can help decision-makers to discover valuable knowledge or influential nodes in a real-world network.
\end{abstract}

\begin{IEEEkeywords}
Critical node problem, Feature Importance, Deep reinforcement learning, Graph attention network.
\end{IEEEkeywords}

%
\IEEEpeerreviewmaketitle


\section{Introduction}
\label{Sec:Introduction}

Networks (also known as graphs) are a general and flexible representation for describing complex systems of interacting entities, which have been widely used to represent real-world data. For example, an air transport network is a complex infrastructure system consisting of various nodes (airports) and links (airport connections). In recent years, graph neural networks (GNNs) have achieved great success in many challenging graph tasks \cite{Li2018,Fan2020,Bai2021,Wang2022,Luo2023}. Among such tasks, detecting a set of critical nodes in sparse networks \cite{Arulselvan2009,Zhou2019,Xu2019,Fan2020,Zhou2021} is an important yet particularly hard task.

Critical node detection is useful in many practical applications \cite{Xu2019,Zhang2020,Doostmohammadian2020,Fan2020,Zhou2021c,Zhou2023b,Zhou2024b,Zhou2024}, where valuable knowledge may be discovered by investigating the identified critical nodes. For example, road networks are extremely vulnerable to cascading failure caused by traffic accidents or anomalous events. Therefore, accurate identification of critical nodes, whose failure may cause a dramatic reduction in the road network transmission efficiency, is of great significance to traffic management and control schemes \cite{Xu2019}. Considering the worldwide air transportation network as a weighted network, many airport connections could be completely or partially disrupted during such extreme events. It is important to identify the most critical airport connections and their impact on network connectivity \cite{Zhou2021c}. According to specific scenarios, different connectivity measures are designed to quantify network functionality appropriately. In this study, we focus on a basic critical node detection problem known as the critical node problem (CNP) \cite{Zhou2021}. It aims to find a subset of nodes such that the removal of these vertices minimizes the pairwise connectivity of the residual network.

Detecting critical nodes in sparse networks is a very challenging task \cite{Arulselvan2009,Baggio2021,Zhou2021b,Salemi2022,Zhou2023b,Zhou2023}. On the one hand, critical nodes are task-specific, i.e., the number and location of critical nodes may vary from task to task. On the other hand, they need to be found at the global scale. To the best of our knowledge, there is no existing deep learning method that can identify task-dependent critical nodes at the global scale in an efficient manner.

Due to its NP-hard nature, state-of-the-art CNP solutions are based on heuristic approaches \cite{Pullan2015,Aringhieri2016,Zhou2019,Zhou2021,Zhou2024}. Domain knowledge and trial-and-error are necessary to design them. They need to be revised or re-executed whenever a change of the problem occurs. It thus becomes extremely expensive to apply them to detect critical nodes in real-world networks that are constantly changing. Moreover, real-world networks from different fields vary significantly in size. For example, transportation networks can be constructed with dozens of nodes to billions of nodes. Although GNNs can operate on networks regardless of their size, can they generalize to networks of sizes that were not seen during their training phase? This question remains unanswered for CNP to the best of the authors' knowledge. It would be highly valuable to design a deep learning method that can train on small networks and generalize to larger networks.

To answer the above-raised question and motivated by the necessity for a scalable and efficient approach that can generalize across different network structures. We aim to learn general representations for various networks and leverage them to make decisions on networks. To this end, we propose a \textbf{F}eature importance-aware \textbf{G}raph attention network and \textbf{D}ueling \textbf{D}ouble-deep-Q-network (FGDD) combined approach to solve CNP. It consists of two phases. In the first phase, we develop a feature importance-aware graph attention network (GAT) based on feature aggregation and a variance-constrained GAT module that learns the importance representation of the nodes. In the second phase, we employ a dueling double deep Q-network (DDQN) as our agent to perform node selection. Our dueling DDQN agent treats node representation from the first phase as a graph state and determines the actions through the Q-network. This work intends to make the following new contributions:
\begin{itemize}
    \item We formulate the process of identifying critical nodes in a network as a sequential decision problem via a Markov decision process (MDP). To solve it, we propose an end-to-end algorithm called FGDD that well combines feature importance-aware GAT with dueling DDQN.
    \item To learn a representation mapping function with robust generalization capabilities for various networks, we aggregate multiple network metrics such as node degree, degree centrality, eigenvector centrality, and PageRank as the foundational features. This composite captures the essential structure information.  Subsequently, we introduce a variance-constrained GAT to maximally utilize these initial features and learn the high-dimension node representations by passing features among nodes.
    \item We experimentally evaluate FGDD on 28 real-world networks. Empirical results show our model not only outperforms two deep reinforcement learning-based algorithms, but also competes well with three heuristic algorithms.
\end{itemize}

The rest of this paper is organized as follows. Section \ref{Sec:Related Work} presents a brief literature review of studies on critical node problems. Section \ref{Sec:Methods} introduces the proposed FGDD approach. Section \ref{Sec:Empirical Results} reports the computational results and comparisons with state-of-the-art algorithms. Finally, Section \ref{Sec:Conclusion} summarizes the work and presents potential research directions.

\section{Related Work}
\label{Sec:Related Work}

\subsection{Exact and Heuristic Methods}
\label{SubSec:Exact and Heuristic Methods}

Detecting critical nodes in sparse networks is proven to be an NP-complete problem \cite{Arulselvan2009}. Considerable efforts have been made to solve it, and a variety of methods have been proposed in the literature \cite{DiSumma2011,Summa2012,Pullan2015,Zhou2021}. Most of existing studies focus on developing exact algorithms and heuristic ones, called traditional algorithm in this paper.

Exact algorithms for CNP are able to guarantee the optimality of their solutions. They are particularly useful for specifically structured graphs such as trees \cite{DiSumma2011} and series-parallel graphs \cite{Shen2012}. For general graphs, CNP has been solved by the branch and cut methods, including a general mixed integer linear programming model \cite{Arulselvan2009}, and a very large model relying on constraint separation \cite{Summa2012}. However, they are very time-consuming when tackling large instances.

Compared to exact algorithms, heuristic algorithms lack theoretical guarantees for the solution quality but are more favorable in practice since they can usually find high-quality solutions in reasonable computational time for large-scale CNP. Existing ones can be divided into local search methods \cite{Pullan2015,Aringhieri2016} and population-based methods \cite{Zhou2019,Zhou2021}. However, domain knowledge and trial-and-error are necessary when designing them, thus requiring considerable efforts and time.

\subsection{Learning-based Methods}
\label{SubSec:Learning-based Methods}

Traditional algorithms suffer from specific limitations when applied to challenging practical tasks. Recent advances in deep learning have shown promising results in solving NP-hard combinatorial optimization problems \cite{Kurin2020,Fan2020,Barrett2020,Wang2021,kou2023identify,jia2023srfa}. They demonstrate appealing advantages over traditional algorithms in terms of computational complexity and scalability \cite{Bengio2021}.

Inspired by the application of graph convolutional networks (GCNs) \cite{Kipf2017,Chen2020,Jepsen2022,Luo2023}, Yu et al. \cite{Yu2020} converted the CNP as a regression problem, and proposed an iterative method to rank nodes by using adjacency matrices of networks and GCNs. Fan et al. \cite{Fan2020} proposed a generic and scalable deep reinforcement learning (DRL) framework named FINDER to find key players via GraphSAGE \cite{Hamilton2017} in complex networks. Kamarthi et al. \cite{Kamarthi2020} developed a deep Q-network (DQN)-based method called geometric-DQN to use structural properties of the available social networks to learn effective policies for the network discovery problem to maximize influence on undiscovered social networks. Zhao et al. \cite{Zhao2020} designed a deep learning model called InfGCN to identify the most influential nodes in a complex network based on GCNs. Note that most of the existing studies focus on identifying nodes in static networks. To predict critical nodes in temporal networks, Yu et al. \cite{Yu2023} proposed a dynamic GCN by combining a special GCN and a long short-term memory network. In addition, graph learning methods are widely used in related areas, like \cite{Yang2023}. The studies mentioned above have greatly advanced the field, but can they be further improved? This work aims to answer this question.

\section{Proposed Method}
\label{Sec:Methods}

In this section, we present our FGDD approach to CNP. We first present CNP and its MDP formulation, and then describe the details of FGDD for solving it.

\subsection{Critical Node Problem}
\label{SubSec:Critical Node Problem}

Given an undirected graph $G = (V,E)$ with $|V|=n$ nodes and $|E|=m$ edges, critical node detection problems (CNDPs) aim to identify a subset of nodes $S \subseteq V$ such that the removal of these nodes in $S$ maximally degrades network connectivity of the residual graph denoted as $G[V\backslash S]$ according to some predefined connectivity metrics. A critical node problem (CNP) is a well-known CNDP \cite{Arulselvan2009}. It aims to find a set of nodes $S=\{v_1,\ldots,v_K\}$ to be removed from $G$, which minimizes the pairwise connectivity of the residual graph $G[V\backslash S]$. The residual graph $G[V\backslash S]$ can be represented by a set of disjoint connected components, i.e., $G[V \setminus S]=\{C_1,C_2,\ldots,C_T\}$, where $C_i$ is the $i$-th connected component in $G[V\backslash S]$. Therefore, its objective function $f(S)$ is defined as follows:
\begin{equation} \label{Equ:Objective Function}
    f(S) = \sum_{i=1}^T\frac{|C_i|(|C_i-1|)}{2}
\end{equation}
where $T$ is the number of connected components, and $|C_i|$ is the size of the connected component $C_i$. Note that $f(S)$ essentially calculates the total number of node pairs still connected by a path in the residual graph $G[V\backslash S]$.

\subsection{Markov Decision Process Formulation of CNP}
\label{SubSec:Markov Decision Process Formulation of CNP}

In this section, we follow the approach presented in \cite{Fan2020} to formulate the critical node problem (CNP) as a Markov decision process (MDP).
\begin{itemize}
    \item \textbf{Environment} ($G_{\text{env}}$) represents the network that is being analyzed.
    \item \textbf{State} ($s_t$)  is defined as the representation of the residual network at time step $t$. Each state $s_t$ captures the hidden information of the residual network at that particular time step by concatenating feature vectors of all nodes generated by a node representation algorithm. The end state is reached when the proportion of removed nodes reaches a certain value.
    \item \textbf{Action} ($a_t$) is the removal of a node $v$ from the current residual network $G_t$.
    \item \textbf{Reward} ($r(a_t,t)$) is the decrease of the pairwise connectivity after taking the action. The optimization objective is to minimize the pairwise connectivity of the residual network. Therefore, the decrease in pairwise connectivity between $s_t$ and $s_{t+1}$ is set as the step-reward after removing a node $v$. Generally, the reward function is written as $r(a_t,t) = f(s_t)-f(s_{t+1})$, where $s_t$ and $s_{t+1}$ are the states of residual networks before and after taking action $a_t$. In this paper, we adopt a new reward function as follows.
        \begin{equation}\label{Equ:Reward Function 1}
            r(a_t,t) = f(s_0)-f(s_{t+1})
        \end{equation}
        where $s_0$ is the initial state, and $s_{t+1}$ is the state of the residual network after taking action $a_t$. The intuition behind (\ref{Equ:Reward Function 1}) is that with the removal of nodes from a network, the remaining nodes have more potential to become critical nodes. We leverage the new reward function to associate the current step with the global process. As shown in Fig. \ref{fig:removal_reward}, the decreased pairwise connectivity of a synthetic network between step 1 and step 2 is $6-{3}=3$, while the one between step 2 and step 3 is ${3}-0=3$. However, it is unfair for the transition to receive a greater reward as the selected node $v_{t=1}$ has the same feature as other nodes in the network while the selected node $v_{t=2}$ is obviously more critical. When we consider the difference between the initial state and the current state, we would have a reward value $6-0=0$, which is greater than the standard version. In addition, the removal of a node in a synthetic network can affect its heterogeneity. Along with the removal, the residual networks can better fit the real networks without a specific generation pattern. Therefore, intuitively, we design the reward considering the whole transition.
        \begin{figure}
        \centering
        \includegraphics[width=1.0\linewidth]{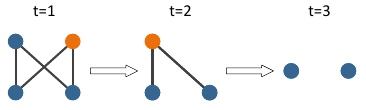}
        \caption{Sequential disintegration of a network, highlighting the impact of node removal over time.}
        \label{fig:removal_reward}
        \end{figure}
\end{itemize}

During a reinforcement learning (RL) process, FGDD collects the trial-and-error samples to update its parameters and becomes increasingly intelligent in solving CNP.

\subsection{Proposed Model}
\label{SubSec:Proposed Model}

\begin{figure*}[!ht]
\centering
\includegraphics[width=2.0\columnwidth]{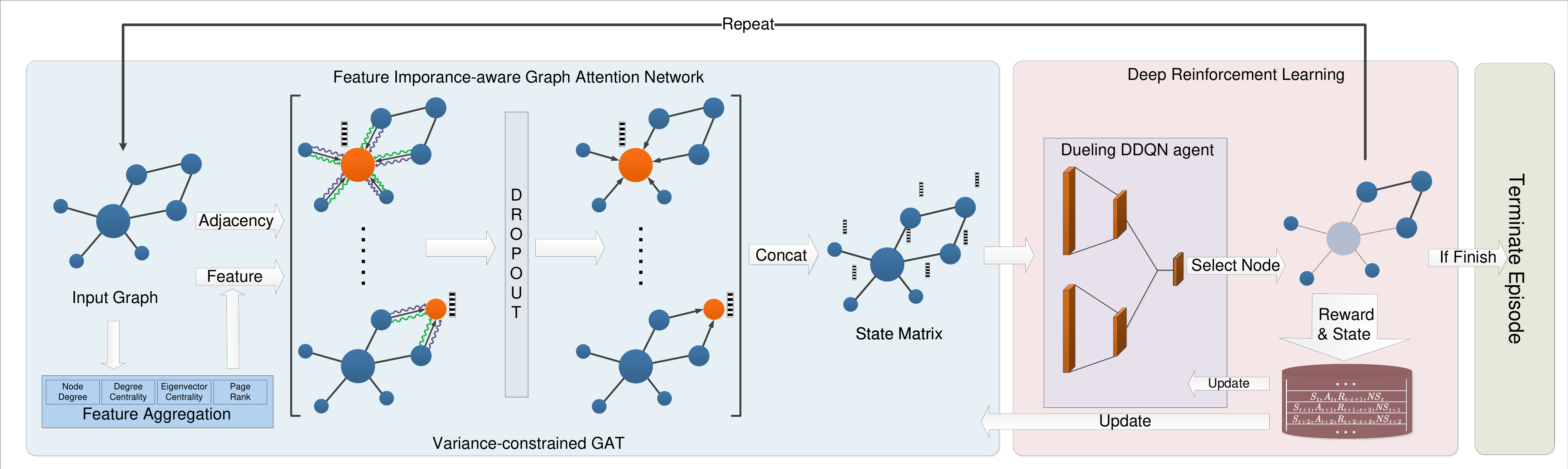}
\caption{Overview of the FGDD framework. We first take basic metrics for a network as initial features for our graph neural network. These metrics provide a fundamental understanding of the network, especially for synthetic networks without given features. These initial features are concatenated and fed into GAT, which employs self-attention mechanisms that allow nodes to weigh the importance of their neighbors' features. We then have representation in latent space for each node and feed these representations to a reinforcement learning agent. After selecting and removing a certain node, we update the replay memory with state, action, and reward.}
\label{Fig:FGDD Framework}
\end{figure*}

To solve CNP, we propose FGDD that combines feature importance-aware GAT and dueling DDQN. As shown in Fig.~\ref{Fig:FGDD Framework}, it is composed of two phases: the graph-feature representation and the feature-based node selection. The first phase aims to present the state of the entire graph, while the second phase employs these representations to form a score function that determines the action for a state. More specifically, node representation is first obtained by a feature importance-aware GAT. The dueling DDQN agent then selects a node to remove from the current network based on the representation information and iteratively constructs a solution to the CNP.

\subsubsection{Feature Representation Phase}
\label{SubSubSec:Encoding Phase}

Traditional algorithms \cite{Arulselvan2009,Pullan2015,Zhou2021} usually operate on a testing network and optimize CNP directly. Unlike them, deep learning methods \cite{Fan2020,Yu2020} aim to extract the features of nodes from a large number of networks, and use a well-trained end-to-end model to predict networks. Therefore, how to represent the contribution of each node to the whole network must be considered.

Compared with task-specific network embeddings, graph neural networks (GNNs) \cite{Wu2020,Huang2021} are considered a powerful method to represent graphs. Inspired by the success of GNNs in both inductive \cite{Chen2020,Hamilton2017,Li2021} and transductive \cite{Kipf2017} problems, we employ a variance-constrained GAT \cite{velickovic2018} cooperated with a mixture of node’s importance metrics (obtained by feature aggregation) to develop a feature importance-aware GAT as our representation function $h$.

Initially, as there are no specific datasets for CNP, we used a large number of synthetic networks to train a general representation function $h$ and apply it to real-world networks. However, synthetic networks are randomly generated without initial features. So, to learn the importance of each node, our node feature is initialized by a matrix $B^{m \times n} $.  Here, $m$ represents the number of nodes, while $n$ captures a set of topological properties. We leverage the node degree as the primary property and three metrics (i.e., degree centrality, eigenvector centrality, and PageRank score) as auxiliary features to describe the network. We aggregate them in the matrix $B^{m \times n}$ as initial features to describe the network's basic information.

Upon initializing the feature matrix $B^{m \times n}$, we apply the GAT to further process these features by using the attention mechanism to capture the relationships between nodes. Within the GAT framework, attention coefficients for nodes are determined as:
\begin{equation}
    \alpha_{ij} = \text{{softmax}}_j (\mathbf{E}^\top [\mathbf{W}B_i \parallel \mathbf{W}B_j])
\end{equation}

Here, $B_i$ is the feature of node $v_i$ extracted from $B$. $\mathbf{W}$ is a weight parameter for linear transformation and $\mathbf{E}$ signifies the parameters of the attention mechanism. $\parallel$ is the concatenation operation. Using these coefficients, we have the new aggregated representations $g(v_i)$ from its neighbors $\mathcal{N}(i)$:
\begin{equation}
    g(v_i) = \sigma \left( \sum_{j \in \mathcal{N}(i)} \alpha_{ij} \mathbf{W} B_j \right)
\end{equation}

To improve the model's performance and robustness, multi-head attention\cite{Zhang2020b} is used to aggregate features from multiple attention heads. Multi-head attention involves multiple sets of the attention mechanism's parameters. The final feature representation $h(v_i)$ in multi-head attention is the concatenation of outputs from each attention head:
\begin{equation}
 \label{eq:feat_agg}
    h(v_i) = \parallel ^K_{k=1} g_k(v_i)
\end{equation}

Where $K$ is the number of attention heads, and $g_k(v_i) $ is the output of the $k$-th attention head for node $v_i$.

In FGDD, due to limited initial features $B^{m \times n}$, we restrict the random dropout mechanism in the original GAT to ensure that more of the available features are retained during each forward pass, maximizing the use of the limited information available. If randomness is high, it could result in inconsistent representations during different forward passes, as crucial nodes or features might get dropped randomly. Such variance from the state in MDP is not needed by the general RL \cite{Anschel2017}. Therefore, compared with the original implementation of GAT, we remove the final dropout layer to preserve the whole input features and set a lower probability for the dropout in the hidden layer.

\subsubsection{Node Selection Phase}
\label{SubSubSec:Decoding Phase}

To select nodes from discrete space, we resort to a deep Q-network (DQN) \cite{Mnih2015} to estimate the best action with the maximum Q value. After empirically evaluating several variants of DQN, we adopt the dueling double deep Q-network (DDQN) \cite{Wang2016} architecture as our node selection method to solve the issue of the overestimation of Q-values in DQN. The dueling DDQN architecture has two separate streams for estimating the state-value and the advantage of each action. These estimates are then combined to obtain the Q-value for each action. This architecture has been shown to be more stable and efficient than other DQN variants.

In our FGDD framework, we use the node representation matrix $\mathbf{H}$ as a state $s$, where each row corresponds to $h(v_i)$ for a node $v_i$. Therefore, in dueling DDQN, we have a Q-value to estimate the importance of each node. To calculate the Q-value, the Q-function is calculated as follows:
\begin{equation}
 \label{eq:node_q}
Q(s,a;\theta_v, \theta_a) = V(s;\theta_v) + A(s,a;\theta_a) - \frac{1}{|\mathcal{A}|}\sum_{a' \in \mathcal{A}}{A(s,a';\theta_a)}
\end{equation}
where $a$ is an action that is a specific node in the graph. $V(\mathbf{F};\theta_v)$ is the value calculated for the current state by a parameter $\theta_v$. And  $A(s,a;\theta)$ is the advantage function, indicating how taking an action  (removing a node in our task) from the action set $\mathcal{A}$ (all nodes in the current graph) compares to the average action for that state. This function uses parameters $\theta_a$. We select the node with the maximum Q-function value as the critical node for the current state and remove it.

\subsubsection{Model Training and Deployment}
\label{SubSec:Model Training and Deployment}

In the reinforcement learning phase of FGDD, we first randomly generated a synthetic network for each epoch. In contrast to real-world networks with a predefined number of nodes $K$ to be removed, synthetic networks lack a predetermined $K$. Therefore, we terminate the training process once a certain proportion $\tau$ of nodes is removed. Our training approach, following \cite{Kamarthi2020}, integrates a target network, experience replay buffer, and epsilon-greedy strategy. The target network serves as a stabilizing factor that is a copy of the main network used for predicting Q-values, but its parameters are only periodically updated to ensure consistent training targets. The experience replay buffer stores the agent's experiences in the form of tuples $(s,a,r,s')$ representing the current state, action taken, resulting reward, and next state. It is then randomly sampled during training ensuring varied learning experiences and efficiency. The epsilon-greedy method fosters a balance between exploration and exploitation. Initially, the agent randomly performs an action with probability $\epsilon$ or performs an action with the highest Q-value with probability $1-\epsilon$. During training, the value of $\epsilon$ is annealed over time to gradually shift from exploration to exploitation.

To accumulate initial experience, the agent selects random nodes from the residual network with a decaying probability $\epsilon$ at the beginning of the training phase using the epsilon-greedy method. At each step, the network uses the adjacency matrix $\mathcal{R}{\emph{adj}}$ and the aggregated feature matrix $B$ to calculate representation for each node using Eq.~\eqref{eq:feat_agg}. The agent then selects a node to remove based on the representation of the current residual network using Eq.~\eqref{eq:node_q}. We have the reward for the agent using Eq.~\eqref{Equ:Reward Function 1}, which compares initial state $s_0$ without removing any node and the current state $s$.

To simplify the training process, we treat the network at each step as a distinct network, and the selected node is directly removed from the network. After receiving rewards from the environment, the agent updates the next state using our feature importance-aware GAT before storing the transition experience into a replay buffer. The main training steps are presented in Algorithm \ref{Alg:Train Network}.

Once our model is well-trained, we evaluate its performance on real-world networks. During the inference phase, given a real-world network with a fixed number of nodes to be removed $K$, FGDD repeatedly estimates the most critical node by Q-function and removes it until $K$ nodes are removed.  To further accelerate the inference, we would remove a batch of nodes in one step for large-scale networks.

\begin{algorithm}[!ht]
\small
\caption{Model Training}
\label{Alg:Train Network}
    \KwIn{Initialize learn rate $\lambda$ and random factor $\epsilon$; memory replay buffer $\mathcal{D}$; current Q network with $\theta_v,\theta_a$; target Q networks with $\bar{\theta_v}=\theta_v,\bar{\theta_a}=\theta_a$; GAT model with parameters $W,E$; the number of epoches $N$}
    \For{episode~$=1$ to $N$}
    {
        Generate training graph $\emph{G}_{\emph{train}}$ and the number of nodes to be removed $K$;\\
        Perform $\epsilon$ decay;\\
        $\emph{G}_{\emph{env}}^{t=0} \gets \emph{sample}(\emph{G}_{\emph{train}})$\\
        $\emph{B}_{\emph{epi}}^{t=0} \gets \emph{FeatureAggregation}(\emph{G}_{\emph{env}}^{t=0})$\\
        $\emph{$\mathcal{H}$}_{\emph{epi}}^{t=0} \gets$ GAT$(B_{\emph{epi}}^{t=0}$,$\mathcal{R}_{\emph{adj}}^{t=0}; W,E)$\\
        \While{not done training}
        {
            Select a random node $v_t $ with probability $\epsilon$ else select node $v_t \gets \emph{argmax Q}(\emph{$\mathcal{H}$}_{\emph{epi}}^{t},v;\theta_v,\theta_a)$;\\
            Remove $v_t$ and observe new graph $\emph{G}_{\emph{env}}^{t+1}$;\\
            Receive reward $r_t$;\\
            Update $\emph{B}_{\emph{epi}}^{t+1} \gets \emph{FeatureAggregation}(\emph{G}_{\emph{env}}^{t+1})$;\\
            Update state $\emph{$\mathcal{H}$}_{\emph{epi}}^{t+1} \gets$ GAT$(\emph{B}_{\emph{epi}}^{t+1}$,$\mathcal{R}_{\emph{adj}}^{t+1}; W,E)$;\\
            Store $\{\emph{$\mathcal{H}$}_{\emph{epi}}^{t},v_t,r_t,\emph{$\mathcal{H}$}_{\emph{epi}}^{t+1}\}$ transition into $\mathcal{D}$;\\
            \If{number of removed nodes~$=K$}
            {
                Terminate current episode;\\
            }
        }
        \If{episode$\mod 20=0$}
        {
            Sample random batch $\delta$ from $\mathcal{D}$;\\
            Update weight $\theta_v,\theta_a, W, E \gets \emph{Adam}(\theta_v,\theta_a, W, E,\lambda,\delta)$;\\
            \If{number of updates$\mod 300=0$}
            {
                $\bar{\theta_v} \gets \theta_v, \bar{\theta_a} \gets \theta_a$;\\
            }
        }
    }
    \KwOut{$\theta_v,\theta_a, W, E$}
\end{algorithm}

\subsection{Complexity Analysis}
\label{SubSec:Complexity Analysis}

The time complexity of FGDD is composed of two parts. Our encoder consists of a feature aggregation and variance-constrained GAT. Our feature aggregation method aggregates four metrics into a matrix that well represents network information. The time complexities of computing both node degree and degree centrality are $O(n)$, where $n$ is the number of nodes. Those of computing eigenvector centrality and PageRank are $O(n\log n)$ and $O(\xi m)$, respectively, where $\xi$ is a constant and $m$ means the number of edges. Therefore, the total time complexity of feature aggregation takes $O(n+n\log n+m)$, or simplified to $O(n\log n+m)$. The complexity of variance-constrained GAT is $O(nll'+ml')$, where $l$ is the number of input features and $l'$ is the number of features computed by a GAT attention head, both of which are small constants. In implementation, we would have multiple attention heads, but the number of attention heads is a predefined constant, so it is not shown in complexity. Therefore, the complexity of the encoding phase is $O(n\log n +nll'+ml')$. We adopt dueling DDQN as our decoder with its complexity being simplified written as $O(n\log n)$ for forward propagation via fully connected layers and max-Q selection. Hence, the total complexity of FGDD is $O(n\log n +nll'+ml')$.

\section{Experiments}
\label{Sec:Empirical Results}

\subsection{Dataset}
\label{SubSec:Dataset}

Our FGDD model is first trained on a large number of small synthetic networks of 50-150 nodes. To avoid overfitting, we always train a network once. Note that these synthetic networks are generated by three widely-used network models \cite{Fan2020,Yu2020,Zhou2021}, i.e., Barab\'{a}si-Albert, Watts-Strogatz, and Erd\H{o}s-R\'{e}yi models. The purpose of using a mixture of synthetic networks is to generalize the model to different types of networks.

To evaluate the performance of our FGDD model, we test FGDD on 21 networks from widely-used CNP benchmarks \cite{Zhou2019,Zhou2021} and 7 networks from the Network Data Repository \cite{Rossi2015}. They are medium and large real-world networks obtained from diverse domains, such as the flight connections between major US airports in 1997, the train network around the city of Rome, the flight network among airports in the European Union in 2014, the network of flight connections in the USA, the electricity distribution network in the USA, and the social network constructed from relations on Facebook.

\subsection{Baseline Methods}
\label{SubSec:Baseline Methods}

We compare FGDD with two categories of baseline methods: traditional construction heuristic methods and DRL-based methods. Construction methods aim to remove nodes from a network in a greedy or random way. As shown in \cite{Fan2020}, CI and HDA greedily remove nodes with the highest collective influence and the highest degree from a network, respectively, while RAND randomly removes nodes from a network. DRL-based methods construct an end-to-end model to remove nodes, including FINDER \cite{Fan2020} and GCN-DRL.

Recently, graph convolutional networks (GCNs) have been successfully applied to predict influential nodes in complex networks. For example, Yu et al. \cite{Yu2020} ranked the nodes by their scores and selected nodes greedily via GCN. Zhao et al. \cite{Zhao2020} developed InfGCN consisting of a GCN and three fully connected layers, which is similar to a combination of GCN and DQN, and used it to select top influential nodes. Note that the existing two GCN-based methods are not publicly available. Therefore, we merge these two GCN-based methods and develop a new baseline method, i.e., graph convolutional network based deep reinforcement learning (GCN-DRL). GCN-DRL shares the same network architecture as InfGCN. The main difference is that GCN-DRL treats the three fully-connected layers as a Q-network, and updates the parameters by RL as well as FINDER. In particular, GCN is used to compute the graph representation, and three fully-connected layers are employed as DRL agent to select critical nodes. For other baseline methods, we adopt the best model released by FINDER's authors to test without any change. We also use the same implementation of CI and HDA \cite{Fan2020}.

We present the time complexities of FGDD and five baseline methods in Table \ref{Tab:Complexity Comparison}. From it, we observe that FGDD is highly competitive in comparison with state-of-the-art deep learning methods. RAND is the fastest but worst in performance to be shown next.

\begin{table}[!ht]
\small
\centering
\caption{Comparison of the time complexity}
\label{Tab:Complexity Comparison}
\setlength{\tabcolsep}{5mm}{
\begin{tabular}{lr}
\toprule
Method  & Complexity\\
\midrule
FGDD    & $O(n\log n+nll'+ml')$ \\
FINDER  & $O(n\log n + n + E)$ \\
GCN-DRL & $O(n\log n+nl^2)$ \\
CI      & $O(n^2\log n)$ \\
HDA     & $O(n\log n)$ \\
RAND    & $O(1)$ \\
\bottomrule
\end{tabular}}
\end{table}

\subsection{Experiment Details}
\label{SubSec:Hyperparameter Settings}

We implement the code in PyTorch and Networkx. All experiments are run on the server with Intel E5-2690 CPUs and two NVIDIA Tesla P100 GPUs. For feature representation phase, we develop the GAT model with eight attention heads, each head has two layers. The hidden and output feature dimensions are set to 60 and 96, respectively. We only deploy the dropout layer with rate of 0.3 between two layers to constrain the variance. 8 attention heads are used to learn the representation. For the node selection phase, we adopt typical dueling DDQN as the agent. Both value and advantage networks are implemented as two linear layers with embedding dimensions being 256 linking with ReLU activation function.

We use the Adam optimizer \cite{Adam2015} to train dueling DDQN and feature importance-aware GAT with the learning rates of 0.001 and 0.005, respectively. We use the mean square loss function to learn from the reward. The eval network is trained per 20 epochs, and the update frequency of a target network is 300. In each epoch, a synthetic network is generated to train the model.

\subsection{Computational Results}
\label{SubSec:Computational Results}

To evaluate FGDD against baseline methods, we conduct three groups of experiments on real-world networks. First, we report their results on 28 real-world networks. Second, we improve their results by a simple local search. Finally, we demonstrate their performance curves under different $K$ values.

\subsubsection{Results without local search}
\label{SubSubSec:Results Without Local Search}

Table \ref{Tab:Comparison of FGDD With Baseline Methods} summarizes the results between our FGDD and five baseline methods on 28 real-world networks, where the best results of each instance are in bold. In addition, we give the number of instances on which FGDD obtains better (Wins), equal (Ties), and worse (Loses) results than the corresponding baseline algorithms. At its bottom, we also give the $p$-values of the Wilcoxon signed ranks test recommended in \cite{Demsar2006}.

\begin{table*}[!htb]
\centering
\caption{Comparative results of FGDD and baseline methods on real-world networks in terms of pairwise connectivity.}
\label{Tab:Comparison of FGDD With Baseline Methods}
\begin{tabular}{lrrr|rrrrrr}
\toprule
Instance & \#Nodes & \#Edges & $K$ &\textbf{FGDD} & FINDER & GCN-DRL & CI & HDA & RAND \\
\midrule
Bovine&121&190&3        & \textbf{268} &\textbf{268} &\textbf{268} &\textbf{268} &\textbf{268} &6903\\
Enron email&143&623&14     & \textbf{7750} & 7875 & 8128 & 7876 & 8128 & 8078 \\
Circuit&252&399&25      & 24090 & 11681 & \textbf{9874} & 11414 & 11681  & 24626  \\
TreniRoma&255&272&26    & 1836 & 2770 & 2501 & 5655 & \textbf{1731} & 3775 \\
Ecoli&328&456&15& 1668  & \textbf{1579} & 1911 & 2137 & 1668 & 44502 \\
USAir97&332&2126&33     & \textbf{15764} & 16041 & 23037 & 15944 & 15944 & 41876 \\
HumanDisease&516&1188&52& 9473 & \textbf{1351} & 3094 & 13849 & 1651 & 72280 \\
Caltech36&769&16656&77    & \textbf{209634} & 220786 & 223452 & 222117 & 222117 & 232637 \\
Wiki-Vote&889&2914&89       & 206429 & \textbf{203247} & 234976 &  228182 & 208374 & 298125 \\
Reed98&962&18812&96       & \textbf{344865} & 365940 & 367653 & 366796 & 366796 & 372299 \\
email-univ&1133&5451&113        & \textbf{408161} & 434780 & 448882 & 493522 & 436649 & 504316 \\
EU\_flights&1191&31610&119& \textbf{439460} & 462259 & 537168 & 491543 & 492535 & 545089 \\
Haverford76&1446&59589&145& \textbf{814726} & 840456 & 841753 & 841753 & 841753 & 843571  \\
openflights&1858&13900&186& \textbf{175447} & 182258 & 223573 & 251088 & 193900 & 847617 \\
OClinks&1899&13838&190  & \textbf{743619} & 762026 & 831424 & 825000 & 785662 & 1379317 \\
yeast1&2018&2705&202    & 114829 & 1826 & 1937 & 34468 & \textbf{1778} & 925890 \\
Bowdoin47&2252&84387&225  & \textbf{1943407}& 2025079 & 2037172  & 2035154 & 2029106 & 2044447 \\
facebook&4039&88234&404 & 5400494 & \textbf{3771740} & 6190140 & 4688231 & 5339614 & 6586685 \\
powergrid&4941&6594&494 & 2922557 & 118780 & 193854 & 1036812 & \textbf{76046} & 8320922  \\
Hamilton1000&1000&1998&100   & \textbf{393828} & 400067 & 403651 & 402753 & 402753 & 404190 \\
Hamilton2000&2000&3996&200   & \textbf{1581531} & 1606530 & 1617301 & 1617301 & 1613706 & 1616942 \\
Hamilton3000a&3000&5999&300  & \textbf{3555111} & 3614020 & 3643650 & 3630165 & 3638254 & 3640412 \\
Hamilton3000b&3000&5997&300  & \textbf{3563115} & 3600590 & 3643650 & 3632860 & 3630167 & 3643650 \\
Hamilton3000c&3000&5996&300  & \textbf{3536470} & 3616710 & 3635557 & 3632861 & 3624779 & 3638793 \\
Hamilton3000d&3000&5993&300  & \textbf{3549780} & 3627470 & 3638253 & 3622086 & 3622087 & 3642031 \\
Hamilton3000e&3000&5996&300  & \textbf{3563115} & 3627470 & 3643650 & 3635556 & 3627472 & 3640415 \\
Hamilton4000&4000&7997&400   & \textbf{6253422} & 6442260 & 6467406 & 6453028 & 6438666 & 6471004 \\
Hamilton5000&5000&9999&500   & \textbf{9876790} & 10068800 & 10113753 & 10100265 & 10100265 & 10106561 \\
\midrule[0.5pt]
Wins$\mid$Ties$\mid$Loses&$-$&$-$&$-$&$-$&20$\mid$1$\mid$7&23$\mid$1$\mid$4&23$\mid$1$\mid$4&20$\mid$2$\mid$6&28$\mid$0$\mid$0\\
\midrule[0.5pt]
$p$-value&$-$&$-$&$-$&$-$& \textbf{3.25e-2} &\textbf{2.30e-3} &\textbf{8.80e-3} &\textbf{1.33e-2} &\textbf{3.79e-6} \\
\bottomrule
\end{tabular}
\end{table*}

We observe in Table \ref{Tab:Comparison of FGDD With Baseline Methods} that our FGDD model shows excellent performance. In particular, it finds the best results on 20 out of 28 networks. Our FGDD wins all baseline algorithms on at least 20 of 28 networks. At a significance level of 0.05, it significantly outperforms two DRL-based algorithms (i.e., FINDER and GCN-DRL) and three heuristic algorithms (i.e., CI, HDA, and RAND). Using the aggregated feature to initially describe a model makes FGDD outperforms similar structure methods like FINDER and GCN-DRL.

\subsubsection{Results with local search}
\label{SubSubSec:Results With Local Search}

FGDD can be combined with other existing heuristic search methods to further improve its performance, which is confirmed by using a recently proposed component-based neighborhood search (CBNS) \cite{Zhou2019}. CBNS is a simple and fast local search, which relies on a focused and reduced neighborhood by integrating a large component-based node exchange strategy and a node weighting scheme. In our experiment, the final result of each algorithm is further improved by CBNS with the stopping condition that the given number of iterations without improvement is reached, i.e., $\emph{MaxIters}=1000$. We report the average results over five runs of local search. We summarize their improved results in Table \ref{Tab:Comparison of FGDD With Baseline Methods Enhanced by Local Search}.

\begin{table*}[!htb]
\centering
\caption{Comparative results of FGDD and baseline methods improved by local search on real-world networks in terms of pairwise connectivity.}
\label{Tab:Comparison of FGDD With Baseline Methods Enhanced by Local Search}
\begin{tabular}{lrrr|rrrrrr}
\toprule
Instance &\#Nodes &\#Edges &$K$ &\textbf{FGDD} & FINDER & GCN-DRL & CI & HDA & RAND\\
\midrule
Bovine&121&190&3& \textbf{268} & \textbf{268} & \textbf{268} & \textbf{268} & \textbf{268} & \textbf{268} \\
Enron email&143&623&14     & \textbf{5156} &\textbf{5156} &\textbf{5156} &\textbf{5156} & \textbf{5156} &5265 \\
Circuit&252&399&25& \textbf{2280} & 2366 & 2635 & 2371 & 2604 & 2527 \\
TreniRoma&255&272&26& \textbf{936} & 938 & 947 & 941 & 938 & 947 \\
Ecoli&328&456&15& \textbf{806} & \textbf{806} & \textbf{806} & \textbf{806} & \textbf{806} & 815 \\
USAir97&332&2126&33& \textbf{5444} & 5484 & \textbf{5444} & 5452 & \textbf{5444} & 5452 \\
HumanDisease&516&1188&52& \textbf{1116} & 1117 & 1119 & 1120 & \textbf{1116} & 1121 \\
Caltech36&769&16656&77    & \textbf{189436} & 190064 & 191533 & 192029 & 191660 & 195266 \\
Wiki-Vote&889&2914&89       & \textbf{151780} & 154205 & 167470 & 169933 & 157321 & 168593 \\
Reed98&962&18812&96       & \textbf{328782} & 339244 & 337434 & 338419 & 338913 & 339904 \\
email-univ&1133&5451&113        & 371631 & \textbf{366135} & 373404 & 375782 & 372315 & 379768 \\
EU\_flights&1191&31610&119& 368879 & 366316 & 367342 & 367173 & 369246 & \textbf{365626} \\
Haverford76&1446&59589&145& \textbf{804548} & 824457 & 821894 & 821378 & 825230 & 824974 \\
openflights&1858&13900&186& \textbf{37648} & 38648 & 40260 & 38519 & 39800 & 210305 \\
OClinks&1899&13838&190& \textbf{642929} & 643155 & 700939 & 663533 & 654357 & 897544 \\
yeast1&2018&2705&202& 2015 & \textbf{1428} & 1442 & 1683 & 1433 & 33735 \\
Bowdoin47&2252&84387&225  & \textbf{1924921} & 1989023 & 1990615 & 2000208 & 1994606 & 2007415 \\
facebook&4039&88234&404& 4755478 & \textbf{3515847} & 5571466 & 3644206 & 5076011 & 5786544 \\
powergrid&4941&6594&494& 74766 & 23562 & 326841 & 37859 & \textbf{22050} & 2157729\\
Hamilton1000&1000&1998&100& \textbf{377168} & 386240 & 392236 & 3877662 & 391017 & 391907 \\
Hamilton2000&2000&3996&200& \textbf{1554969} & 1578688 & 1576558 & 1589733 & 1572660 & 1591167 \\
Hamilton3000a&3000&5999&300& \textbf{3494053} & 3571677 & 3601665 & 3579163 & 3595230 & 3596841 \\
Hamilton3000b&3000&5997&300& \textbf{3513109} & 3548200 & 3602736 & 3591477 & 3583437 & 3590938 \\
Hamilton3000c&3000&5996&300& \textbf{3490360} & 3561526 & 3577035 & 3579691 & 3576485 & 3591477 \\
Hamilton3000d&3000&5993&300& \textbf{3500424} & 3563691 & 3593078 & 3565803 & 3576481 & 3600592 \\
Hamilton3000e&3000&5996&300& \textbf{3499888} & 3578618 & 3580767 & 3582904 & 3570615 & 3604888 \\
Hamilton4000&4000&7997&400& \textbf{6253423} & 6356421 & 6412857 & 6378537 & 6368528 & 6401412 \\
Hamilton5000&5000&9999&500& \textbf{9781919} & 9969463 & 10054483 & 10000772 & 10024904 & 10051796 \\
\midrule
Wins$\mid$Ties$\mid$Loses&$-$&$-$&$-$&$-$&20$\mid$3$\mid$5&22$\mid$4$\mid$2&22$\mid$3$\mid$3&21$\mid$5$\mid$2&26$\mid$1$\mid$1\\
\midrule
$p$-value&$-$&$-$&$-$&$-$& \textbf{1.19e-2} &\textbf{5.61e-5} &\textbf{2.70e-3} &\textbf{2.62e-4} &\textbf{1.23e-5} \\
\bottomrule
\end{tabular}
\end{table*}

From Table \ref{Tab:Comparison of FGDD With Baseline Methods Enhanced by Local Search}, we observe that FGDD outperforms baseline methods on most of the real-world networks. In particular, it achieves the best results on 22 out of 28 networks. Compared with two DRL-based methods, it finds better results on at least 21 networks, and equal results on at least 3 networks. Compared to heuristic search methods, FGDD also shows excellent performance, finding better results on at least 22 networks, worse result on at most one instance, and equal results on the remaining instances. At a significant level of 0.05, FGDD is significantly better than all five baseline methods. These observations show that existing heuristic algorithms can also benefit from FGDD.

\subsubsection{Results under different $K$ values}
\label{SubSubSec:Results Under Different $K$ Values}

Note that results reported in Tables \ref{Tab:Comparison of FGDD With Baseline Methods}-\ref{Tab:Comparison of FGDD With Baseline Methods Enhanced by Local Search} are obtained under a fixed $K$ value. That is, the allowable number of removed nodes is determined in advance. To further evaluate its effectiveness, we compare the performance curves of FGDD and five baseline methods under a series of $K$ values on four representative real-world networks: flight network (i.e., EU\_flights), hamiltonian cycles network (i.e., Hamilton2000), social network (i.e., OClinks) and communication network (i.e., email-univ). These real-world networks are obtained from different fields, with node counts ranging from 1000 to 2000.

\begin{figure*}[!htb]
\centering
\includegraphics[width=1.8\columnwidth]{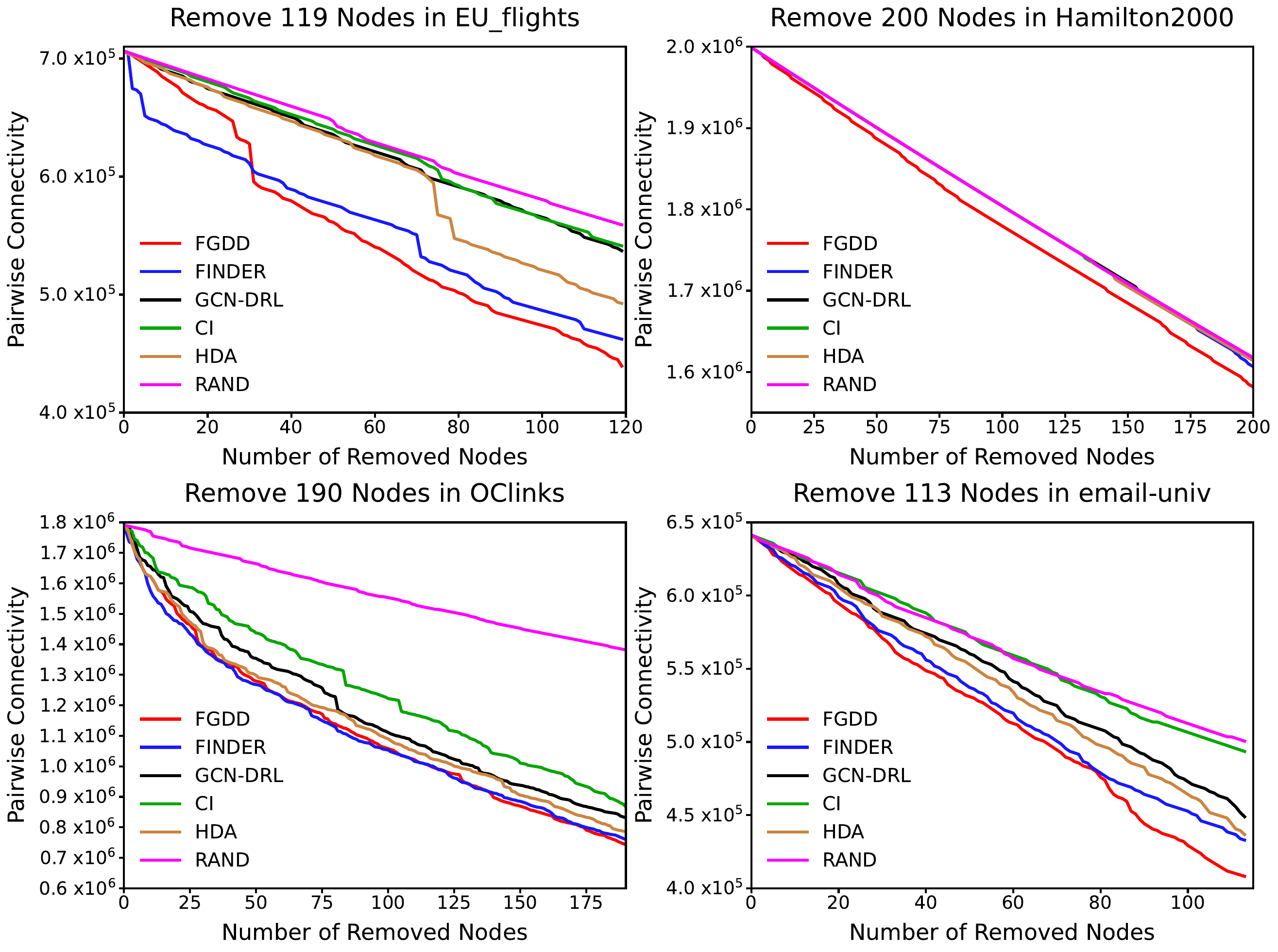}
\caption{Performance curves of FGDD and baseline methods under different $K$ values. Here we show the pairwise connectivity decreasing trend for each step after removing the selected node.}
\label{Fig:Performance Curves With Different K Values}
\end{figure*}

As shown in Fig.~\ref{Fig:Performance Curves With Different K Values}, FGDD consistently outperforms five baseline methods on all four networks, identifying nodes with less pairwise connectivity compared with baselines. This observation further confirms the superiority of FGDD over both DRL-based methods and heuristic methods. Expanding on this, the consistent performance of FGDD can be attributed to its innovative framework and underlying mechanisms. By effectively integrating domain-specific initial features with advanced learning techniques, FGDD enhances its decision-making process at each iteration. This results in a more nuanced and precise node selection strategy, which is beneficial when dealing with complex networks.

\subsection{Ablation Study}
\label{SubSec:Ablation Study}

\begin{figure*}[!htb]
\centering
\includegraphics[width=1.8\columnwidth]{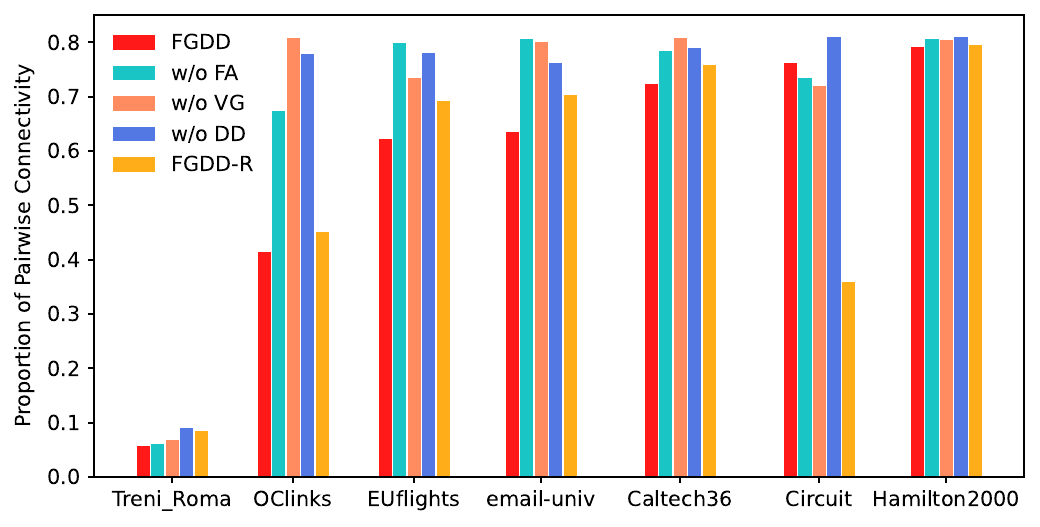}
\caption{Ablation study on real-world networks. We compare our FGDD method with its four variants by removing the feature aggregation (w/o FA) method, variance-constrained GAT (w/o VG) module, and dueling DDQN (w/o DD) module, respectively. In addition, We also report the results of FGDD under a different reward function (FGDD-R for short).}
\label{Fig:Ablation}
\end{figure*}

To investigate the effectiveness of different algorithmic components proposed in our FGDD model, we perform three ablation studies on 7 representative real-world networks. In our ablation studies, we experimentally compare our FGDD model with its four variants. Comparative results are shown in Fig.~\ref{Fig:Ablation}.

To evaluate the effectiveness of the feature aggregation (FA) method used in the feature importance-aware GAT, we examine the same implementation of FGDD, but only use node degree as the initial feature, and exclude the FA component, named as w/o FA in Fig.~\ref{Fig:Ablation}. The node degree is a simplistic metric that captures local information. However, the usage of FA allows for a better representation of the nodes. This enriched representation is instrumental in providing the GAT with a more detailed context to enhance the model training and make the model outperform the version without FA observing from Fig.~\ref{Fig:Ablation}. This suggests that feature aggregation is quite important to model performance.

We study the benefit of the variance-constrained GAT (VG for short) module that introduces modifications to the dropout layer mentioned in the feature representation phase. We replace it with the original GAT (w/o VG) to explore how variance influences the efficiency of FGDD. Here, GAT would randomly drop out initial features and have twice the dropout rate as VG. Comparative results in Fig.~\ref{Fig:Ablation} show that FGDD (with VG) outperforms the model with original GAT (w/o VG), thus verifying the importance of maintaining most of the initial features for CNP to avoid performance decrease, and confirming the effectiveness of our variance-constrained GAT. Moreover, the comparative results involving the dueling deep Q-network (DD) and the basic deep Q-network DDQN (w/o DD) within the FGDD framework highlight the need for a better DRL algorithm to address the CNP problem. The performance of a model with basic DDQN is worse than FGDD (with DD), indicating the significant performance boost introduced by dueling DDQN.

Moreover, we experimentally compare our reward function in Eq.~\eqref{Equ:Reward Function 1} with the basic version (named FGDD-R) that considers the decreased pairwise connectivity between the two steps. In most cases, our reward function is better than the typical formula. It validates the effectiveness of our design of reward function that considering the whole transition of a network is better than only considering two consequent steps.

\begin{table*}[ht!]
\centering
\caption{Performance comparison of FGDD under different GAT parameters in terms of pairwise connectivity}
\label{tab:gatattentionnumber}
\begin{tabular}{cccccccc}
\toprule
$|\mu|$& Circuit & TreniRoma & Caltech36 & email-univ  & EUflights & OClinks  & Hamilton2000\\
\midrule
2 & 25651 & 2149 & 233591 & 497503 & 568712 & 1023176 & 1599366 \\
4 & 25425 & 1851 & 218136 & 495510 & 566581 & 854805  & 1606531 \\
6 & 24310 & 1599 & 214846 & 442272 & 470942 & 747282  & 1519896 \\
8 & 24090 & 1836 & 209634 & 408161 & 439460 & 743619  & 1581531 \\
\bottomrule
\end{tabular}
\end{table*}

Finally, we investigate the influence of different numbers of attention heads $\mu$ on the performance of FGDD. In this experiment, four different $\mu$ values are used. Detailed comparative results of FGDD under different GAT parameters are summarized in Table~\ref{tab:gatattentionnumber}. The results show that more attention heads can better represent a network and help our agent find critical nodes.

\section{Conclusion}
\label{Sec:Conclusion}

This work presents FGDD as a new DRL-based algorithm for a critical node problem that generalizes well to unseen networks size and structures. FGDD combines feature importance-aware GAT with dueling DDQN. Empirical results on 28 real-world networks show our method achieves superior performance on real-world networks. It significantly outperforms both DRL-based methods and heuristic methods. FGDD can start a search from any valid state, opening the door to its combination with other search heuristics, which we have demonstrated by using a recently proposed component-based neighborhood search. In addition, our model performs well on social networks and has great potential to extend to solve other problems in complex networks, e.g., an influence maximization problem.

As future work, two research potential directions should be pursued. A limitation of our approach is that it does not use any supervised information. We plan to employ it to guide the model training so that our model is able to achieve state-of-the-art performance. In addition, we also plan to verify the effectiveness of our model on a road transport network, such as the identification of critical road infrastructure.

\section*{Acknowledgment}
We would like to thank the referees for their useful comments and suggestions.

\ifCLASSOPTIONcaptionsoff
  \newpage
\fi



\bibliographystyle{IEEEtran}
\bibliography{mybibfiles}

\begin{IEEEbiography}[{\includegraphics[width=1in,height=1.25in,clip,keepaspectratio]{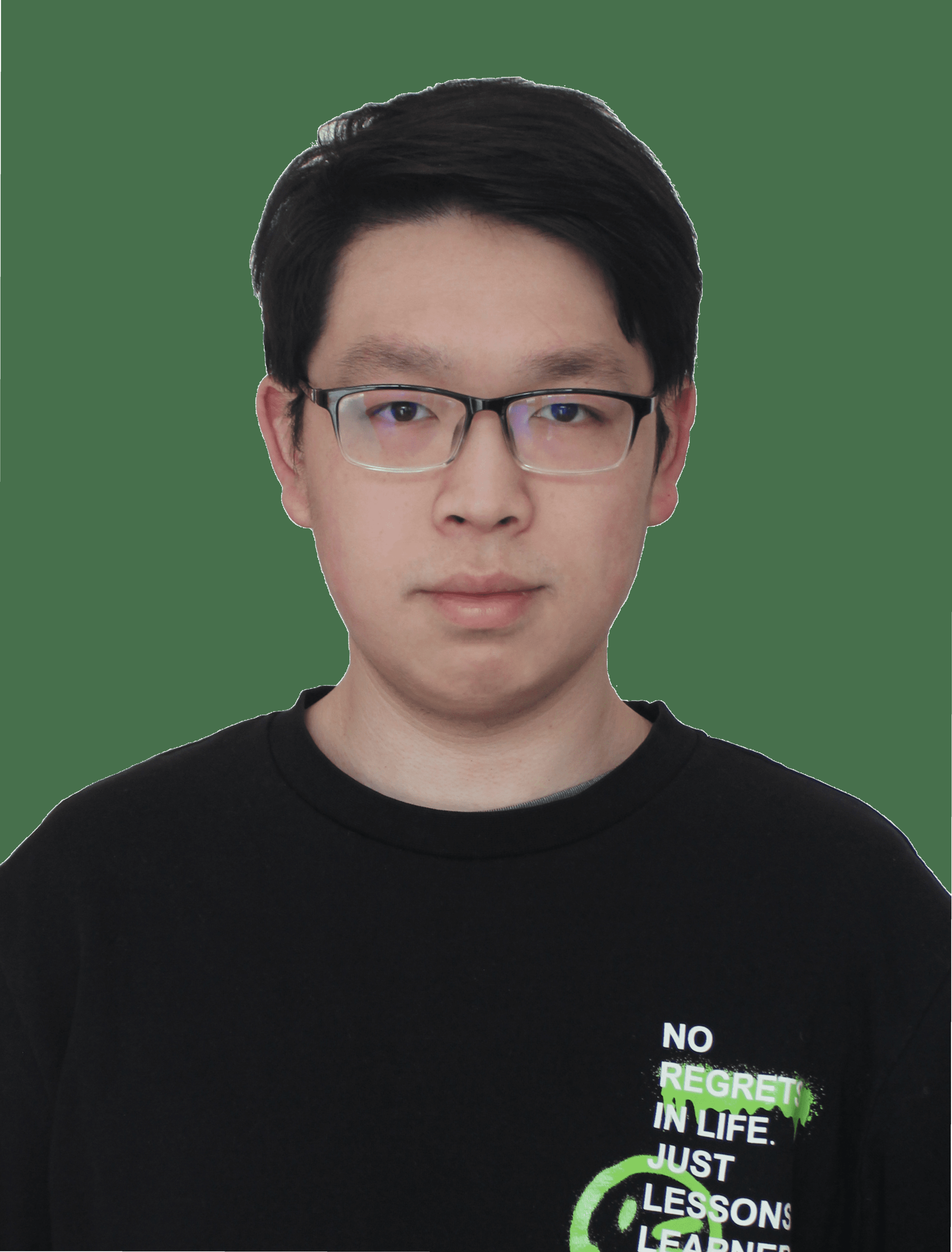}}]{Xuwei Tan} received his B.Eng. degree in Computer Science from the East China University of Science and Technology, Shanghai, China, in 2021. He was a Research Assistant at Shanghai Jiao Tong University. He is now pursuing a Ph.D. degree in Computer Science at the Ohio State University. His research interests include machine learning and combinatorial optimization.
\end{IEEEbiography}

\begin{IEEEbiography}[{\includegraphics[width=1in,height=1.25in,clip,keepaspectratio]{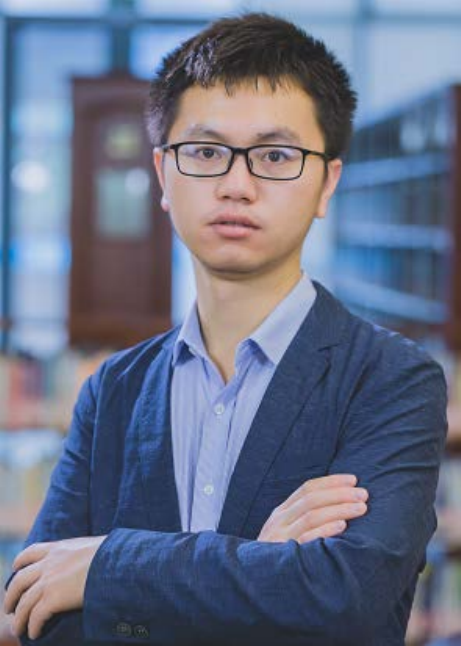}}]{Yangming Zhou} (Senior Member, IEEE) received his Ph.D. degree from the University of Angers, Angers, France, in 2018. He is now an Associate Professor at Shanghai Jiao Tong University, Shanghai, China. His research interests include machine learning, meta-heuristics, combinatorial optimization, and their applications in transportation and logistics. He has published over 50 international journal or conference papers.
\end{IEEEbiography}

\begin{IEEEbiography}[{\includegraphics[width=1in,height=1.25in,clip,keepaspectratio]{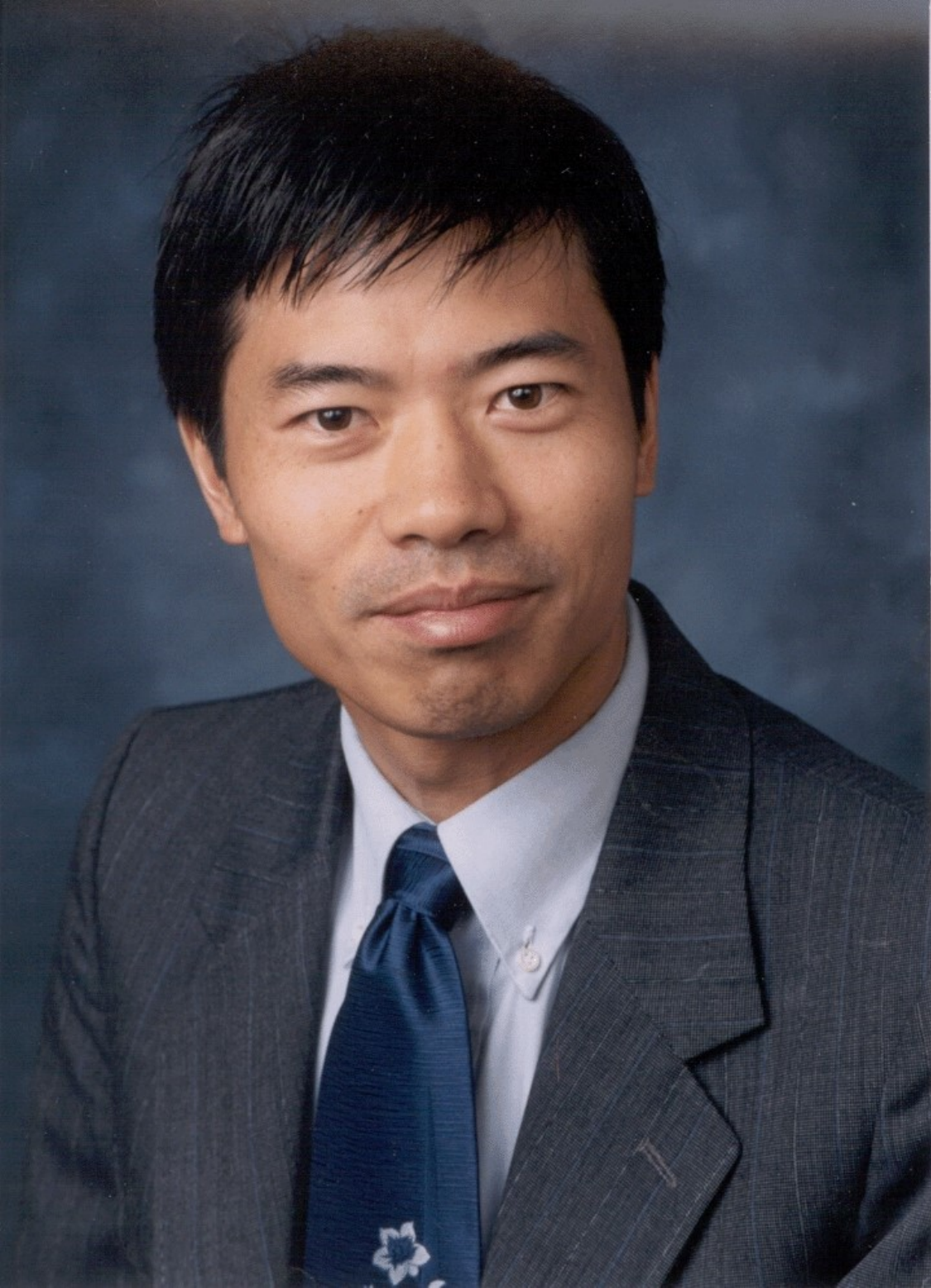}}]{MengChu Zhou} (Fellow, IEEE) received his Ph.D. degree from Rensselaer Polytechnic Institute, Troy, NY in 1990 and then joined New Jersey Institute of Technology where he is now a Distinguished Professor. His interests are in Petri nets, automation, robotics, Internet of Things, big data, and artificial intelligence. He has over 1200 publications including 17 books, 850+ journal papers (650+ in IEEE Transactions), 31 patents and 32 book-chapters. He is Fellow of IFAC, AAAS, CAA and NAI.
\end{IEEEbiography}

\begin{IEEEbiography}[{\includegraphics[width=1in,height=1.25in,clip,keepaspectratio]{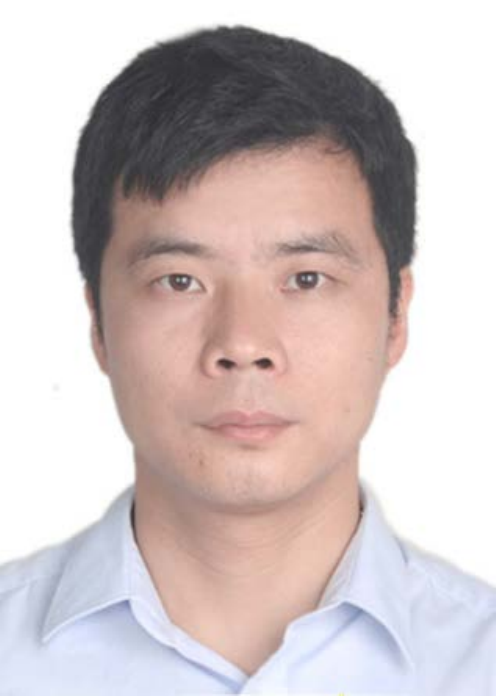}}]{Zhang-Hua Fu} received his Ph.D. degree from Huazhong University of Science and Technology, Wuhan, China, in 2011. He is now a Research Fellow at the Chinese University of Hong Kong, Shenzhen, China. His research interests include combinational optimization, operations research and artificial intelligence. He has over 30 international journal or conference papers, including publications in IEEE TC, IEEE TEVC, IEEE TASE, IEEE TITS, JOC, EJOR, AAAI and IJCAI.
\end{IEEEbiography}

\end{document}